\documentclass{article}

\PassOptionsToPackage{numbers, compress}{natbib}
\usepackage[preprint]{style/neurips_2021}

\usepackage[utf8]{inputenc} 
\usepackage[T1]{fontenc}    
\usepackage{hyperref}       
\usepackage{url}            
\usepackage{booktabs}       
\usepackage{amsfonts}       
\usepackage{nicefrac}       
\usepackage{microtype}      
\usepackage{xcolor}         

\usepackage{wrapfig}
\usepackage{hyperref}
\usepackage{enumitem}
\usepackage{euscript}
\usepackage{threeparttable}
\usepackage{diagbox}
\usepackage{multirow}
\usepackage[subtle]{savetrees}

\usepackage{times}
\usepackage[title]{appendix}
\usepackage{xcolor}

\usepackage{comment}
\usepackage[normalem]{ulem}
\usepackage{mwe}
\usepackage{subfig}
\usepackage{tikz}
\usepackage{lipsum}
\usepackage{amsmath}
\usepackage{amsthm}
\usepackage{txfonts}

\usepackage[linesnumbered, ruled]{algorithm2e}
\SetAlgorithmName{Algo}{Algo}{List of Algorithms}
\SetKwInput{KwInput}{Input}
\SetKwInput{KwOutput}{Output}
\SetKwInput{KwInitialize}{Initialize}
\usepackage{footmisc}

\usepackage{thmtools,thm-restate}

\newcommand\norm[1]{\left\lVert#1\right\rVert}
\usepackage{scalerel,stackengine}
\stackMath
\newcommand\reallywidehat[1]{%
\savestack{\tmpbox}{\stretchto{%
  \scaleto{%
    \scalerel*[\widthof{\ensuremath{#1}}]{\kern.1pt\mathchar"0362\kern.1pt}%
    {\rule{0ex}{\textheight}}
  }{\textheight}%
}{2.4ex}}%
\stackon[-6.9pt]{#1}{\tmpbox}%
}

\title{SHORING: Design Provable Conditional High-Order Interaction Network via Symbolic Testing}

\author{%
  Hui Li\thanks{Hui Li, Xing Fu, Ruofan Wu, Jinyu Xu and Kai Xiao are the key members to conduct the project and theoretical derivations within this paper.  Any questions please contact: lihuiknight@gmail.com}$~~^\ddag$, Xing Fu$^\ddag$, Ruofan Wu$^\ddag$, Jinyu Xu$~~^\ddag$, Kai Xiao$~~^\ddag$, Xiaofu Chang 
  \And Weiqiang Wang\thanks{Weiqiang Wang is the mentor of our project and proposed the key idea of neural symbolic representation.}, Shuai Chen, Leilei Shi, Tao Xiong, Yuan Qi \\
  \\
   Ant Group, Hang Zhou, China 
}

\begin{document}

\maketitle

\begin{abstract}

Deep learning provides a promising way to extract effective representations from raw data in an end-to-end fashion and has proven its effectiveness in various domains such as computer vision, natural language processing, etc. However, in domains such as content/product recommendation and risk management, where sequence of event data is the most used raw data form and experts derived features are more commonly used, deep learning models struggle to dominate the game. In this paper, we propose a symbolic testing framework that helps to answer the question of what kinds of expert-derived features could be learned by a neural network. Inspired by this testing framework, we introduce an efficient architecture named SHORING, which contains two components: \textit{event network} and \textit{sequence network}.
The \textit{event} network learns arbitrarily yet efficiently high-order \textit{event-level} embeddings via a provable reparameterization trick, the \textit{sequence} network aggregates from sequence of \textit{event-level} embeddings. We argue that SHORING is capable of learning certain standard symbolic expressions which the standard multi-head self-attention network fails to learn, and conduct comprehensive experiments and ablation studies on four synthetic datasets and three real-world datasets.
The results show that SHORING empirically outperforms the state-of-the-art methods.

\end{abstract}

\vspace{-4mm}
\section{Introduction}
\label{sec:intro}
\vspace{-2mm}

Effective modeling of feature interactions from sequence data is critical for machine learning models to achieve good performance.
However, it's very time and resource consuming for human experts to use domain knowledge to derive high-order features from multiple sequences of raw event data through manual feature engineering. 
In contrast, deep learning has shown its powerful ability of representation learning in a wide of fields, such as computer vision~\citep{li2013automatic,he2016deep}, natural language processing~\citep{devlin2018bert}, reinforcement learning~\citep{2016+silver+mastering, chen2018neural, chen2019generative}, game theory~\citep{li2019double}, numerical optimization~\citep{sun2020improving} and many other applications~\citep{li2013automatictracking, cheng2016wide,zhao2017rapid, dai2018adversarial, xi2020neural, qu2020intention}.

The data are typically organized as multiple sequences of events.
Given the specific task and processed data, end-to-end deep learning models contain three components:
(1) learn \textit{event-level} representations~(embeddings),
(2) learn \textit{sequence-level} representations from multiple sequences of \textit{event-level} embeddings,
(3) design a learning loss based on the specific task.
To address the first component, efforts have been made toward designing powerful feature interaction methods.
One of the most successful methods is factorization machines (FM). 
Theoretically, FM provides a general framework to learn arbitrary-order feature interaction, however, the computational burden of high-order (i.e., higher than third order) interactions makes second-order FM remaining the dominant practice.
Recently, many FM variants~\citep{guo2017deepfm, lian2018xdeepfm, cheng2020adaptive} are proposed to learn the high-order interaction based on some empirical assumptions.
To address the second component, recurrent neural network~(RNN) provides a flexible framework to learn sequence-level representations.
As one of the most successful RNN variants, long short-term memory~(LSTM)~\citep{1997Hochreiterlstm} achieved strong performance in many areas.
Recently, Transformer~\citep{vaswani2017attention} based on multi-head softmax self-attention beats LSTM and becomes state-of-the-art (SOTA) in a wide field of applications.
For the third component, there are typically many different learning tasks, such as supervised learning, self-supervised learning, contrastive learning and reinforcement learning.
Experts typically design suitable losses for their tasks and use numerical optimization techniques to learn the parameters.

Although there are so many successful end-to-end learning methods, there still remain several challenges in real-world applications, which are listed as follows.

\textbf{$\bullet$ Challenge 1: inductive bias.} 
Most successful neural architectures originate from domains like computer vision and natural language processing, under which precise verification of a fine-grained inductive bias is usually very difficult. Instead, most neural architectures claim their success via attaining competitive test set performance over benchmark datasets. However, in sequence data modeling where the usual goal is to learn some kind of aggregation mechanisms, there is indeed a fairly large design space that one could leverage to formally verify the approximation ability of certain neural architectures. Nevertheless, formal approaches to verifying neural sequence architectures remain largely unexplored.

\textbf{$\bullet$ Challenge 2: high-order interaction.} 
Learning arbitrarily high-order feature interactions for high-dimensional sequence data with limited resources remains a challenging problem for many years~\footnote{For example, the real-time models running on mobile devices typically have a strict computation and memory limit.}.

\textbf{$\bullet$ Challenge 3: unstructured data.} 
The events within multiple sequences data contain various types of unstructured features, such as text, numerical, categorical, and other sparse features. Encoding such unstructured features in a simple yet effective framework will be helpful for the downstream deep learning methods.

In this paper, we provide several solutions for the above challenges.
For challenge 1, we introduce a useful tool named symbolic testing to evaluate the approximation ability for the sequence model based on the kernel two-sample test (see section~\ref{sec:symbolic_testing}).
With the help of symbolic testing, we argue that the standard multi-head self-attention network is not capable of learning certain fundamentally nested symbolic expressions, such as the conditional \textit{count distinct} expressions.
Then we introduce a stronger conditional sequence network following the spirit of symbolic learning.
For challenge 2, we introduce a provable high-order interaction network based on an interesting reparameterization trick to learn arbitrarily high-order feature interaction (see section~\ref{sec:prov_highorder}).
For challenge 3, we introduce a general framework to encode the unstructured event data, including text, categorical and numerical features (see appendix~\ref{app:feature_encoding} because of the limited space). 
At last, we introduce a novel architecture named SHORING~(multiple Sequences High-ORder Interaction Network with conditional Group embedding) to learn the representation for multiple sequences.
We conduct extensive comparisons and ablation studies on synthetic, public and industrial datasets. 
The experimental results show that SHORING empirically beats SOTA.

\vspace{-2mm}
\section{Related Work}
\label{sec:related_work}
\vspace{-2mm}

\begin{wrapfigure}[4]{r}{0.4\textwidth}
\vspace{-16mm}
\centering
    \includegraphics[width=0.37\textwidth]{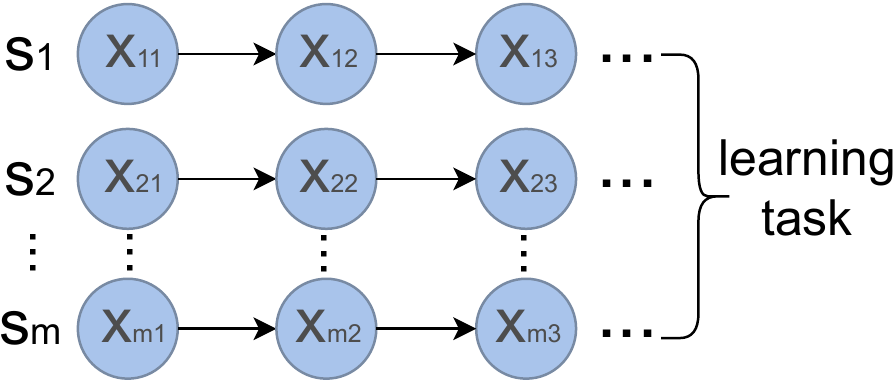}
  \vspace{-2mm}
  \caption{\small $m$ sequences. $x_{ij}$ is $j$-th event in $s_i$. }
  \label{fig:multiple_sequence}
\vspace{-1mm}
\end{wrapfigure}

Learning representation from sequence data is a fundamental problem in real-world applications. 
There are a large number of articles to address this problem.
Wide\&Deep~(W\&D)~\citep{cheng2016wide} is a successful framework with both wide layer and deep layer.
Factorization machine~(FM)~\citep{rendle2010factorization} provides a powerful approach to learn $k$-way variable interactions using factorized parameters under sparsity.
One of the most successful FMs is the second-order version, which requires only linear computation and memory overhead.
To design a tractable high-order interaction network, many researchers~\citep{guo2017deepfm, zhang2016deep, he2017nfm, zhang2016deep, blondel2016higher, blondel2016polynomial} make a lot of efforts based on some heuristic or theoretical approaches.
For example, \citet{xiao2017attentional} proposed an attentional factorization machine~(AFM) and used an attention mechanism to learn the interaction weight between different feature fields.
\citet{wang2017deep} proposed another method named deep and cross network~(DCN) to learn the feature interactions and the degree of cross features increases by one at each cross-layer.
Recently, some experts design hybrid architectures based on high-order feature interaction network and deep sequence networks ($e.g.$, RNN and Transformer variants, $etc$) to learn prediction tasks from sequence data. 
For example, \citep{xi2020neural} uses FM to learn second-order feature interaction and uses a self-attention network to learn the importance of each event.
DSIN~\citep{feng2019deep} uses a GRU with an attentional update gate to model the interest evolving process.
Some other interesting articles, such as MIMN~\citep{pi2019practice} and SIM~\citep{pi2020search}, design novel methods to handle the problem of a long sequence.
Honestly, there are so many articles to learn representation from sequence data, it's hard for us to make a comprehensive survey in the conference paper. We refer readers to these works~\citep{das2017survey, zhou2018deep, choromanski2020rethinking, tay2020long}.

\vspace{-1mm}
\section{Problem and Motivation}
\label{sec:problem}
\vspace{-2mm}

\begin{wrapfigure}[9]{r}{0.4\textwidth}
\vspace{-20mm}
\centering
\includegraphics[width=0.27\textwidth]{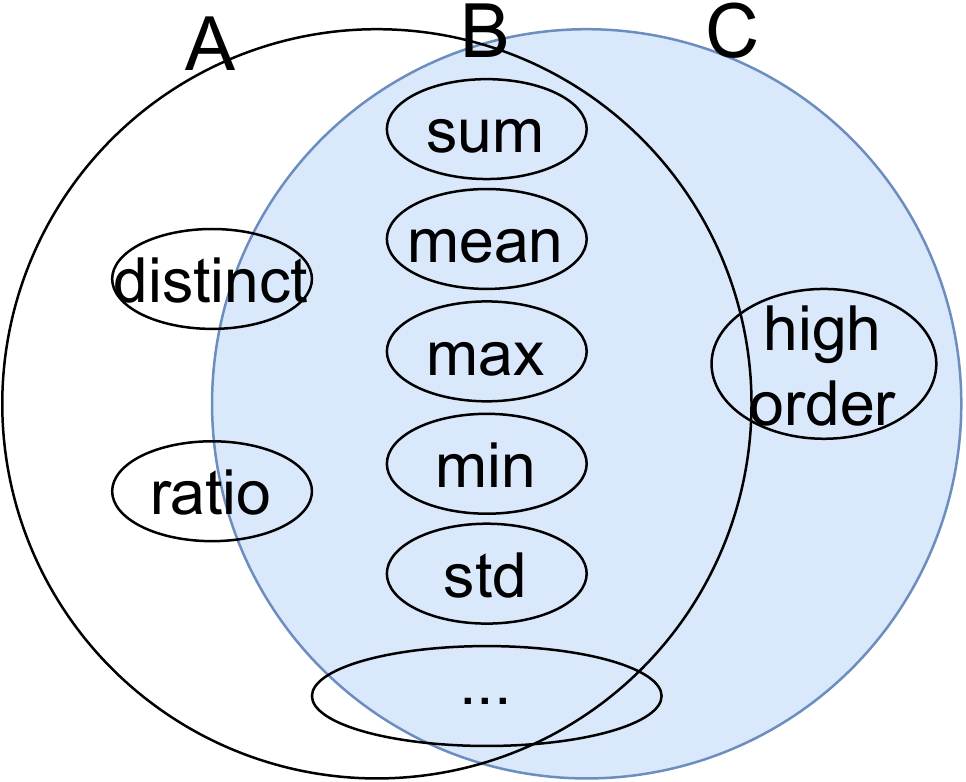}
\vspace{-1mm}
\caption{\small Venn diagram shows advantages of expert-derived features and learning-based features. A and C are expert-derived and learning-based features respectively. B refers to the overlap of A and C.
}
\label{fig:feature_overlap}
\vspace{-1mm}
\end{wrapfigure}

As illustrated in~figure~\ref{fig:multiple_sequence}, given $m$ sequences of events and the specific learning task, we need to design a function approximation model to solve the learning task.
In industrial applications, such as content/product recommendation, risk management, feature engineering typically involves human experts to manually craft numerical features as illustrated by Figure~\ref{fig:feature_overlap} based on their domain knowledge using standard symbolic operators, e.g. \textit{sum}, \textit{count}, \textit{mean}, \textit{standard deviation}, \textit{count distinct} and \textit{ratio} over filtered historical events. With human-derived features, function approximation models are built using machine learning algorithms, such as logistic regression, gradient boosting trees~\citep{chen2016xgboost} and RNN $etc$.
Recently, researchers employed end-to-end neural sequence networks~\citep{xi2020neural}, e.g. standard softmax self-attention network, to learn features from sequence data and demonstrated successes in various scenarios. However, in the aforementioned domains, deep learning models fail to consistently win over manual feature engineering~\citep{jannach2020deep}.
Hence it's quite natural to ask:

\begin{center}
\vspace{-1mm}
  \framebox{
  \parbox{0.9\columnwidth}{ 
  {\bf Question 1:} {Is the standard sequence network, $i.e.$, self-attention neural network, as the SOTA sequence model architecture, capable of learning expert-based features well?}    
  }}
\vspace{-1mm}
\end{center}

\vspace{-1mm}
\section{Symbolic Testing}
\label{sec:symbolic_testing}
\vspace{-2mm}

\vspace{-1mm}
\subsection{Hypothesis Testing for Symbolic Expressions}
\vspace{-1mm}

To answer the above question, it's necessary to establish a framework to test the ability of representation learning for sequence models.
Inspired by the statistical hypothesis testing~\citep{kendall1946advanced}, in this section, we introduce a tool named \textbf{symbolic testing} which can report how well does one neural network learns a given symbolic expression.
The key idea of our symbolic testing is very simple.
Given one symbolic expression and a neural network, we collect $n$ synthetic data $\{(x_i, y_i)\}_{i \in [1, n]}$ generated from this expression under specific data distribution and use the neural network $f(x_i)$ to approximate $y_i$, where $x_i$ is the sequence data and $y_i$ is the return of the given symbolic expression.
We construct a statistical two-sample test to determine if $\{f(x_i)\}_{i \in [1, n]}$ and $\{y_i\}_{i \in [1, n]}$ are drawn from different distributions.
In this paper, we use kernel two-sample test~\citep{gretton2012kernel} and formulate a test statistic based on the following empirical estimate of the \textit{maximum mean discrepancy}~(MMD)
\begin{flalign}
\label{eq:kernel_test}
\vspace{-2mm}
  \begin{aligned}
\reallywidehat{MMD}(P_{f(x)}, P_y) = \norm{ \frac{1}{n} \sum^{n}_{i=1}{\phi(f(x_i))} - \frac{1}{n} \sum^{n}_{i=1}{\phi(y_i))} }_{\mathcal{H}}
  \end{aligned}
\end{flalign}
obtain a two-sample test of the null hypothesis that both samples stem from the same distribution, where $\phi(x)$ is the feature map that generates the reproducing kernel Hilbert space $ \mathcal{H} $, and $ \norm{\cdot}_{\mathcal{H}} $ is the norm induced by its inner product, $P_{f(x)}$ and $P_y$ are the distributions of $f(x)$ and $y$. Refer to \citep{gretton2012kernel} for more details.

\vspace{-1mm}
\subsection{Interesting Observations}
\vspace{-1mm}

According to our experiments, under specific data distribution, the standard softmax self-attention neural network is able to learn many symbolic expressions well, such as those based on \textit{sum}, \textit{mean}, \textit{max}, \textit{min}, $etc$.
However, what it is NOT capable of learning are certain conditional symbolic expressions, such as expressions based on \textit{count distinct} operator.
Let us explain this issue with a simple example.
Given a user's login event sequence, we want to count the number of different cities under some conditions, such as, login time is between 1 am and 5 am.
The above aggregation could be computed in SQL (Structured Query Language) by Algorithm~\ref{alg:count_distinct_1}.

To understand Algorithm~\ref{alg:count_distinct_1}, we decompose \textit{count distinct} expression into an equivalent form which consists two consecutive symbolic expressions. 
(1) \textit{group-by}: compute cities' frequency within the sequence under the specific condition;
(2) \textit{conditional sum}: count the number of cities with positive frequency.
We introduce the transformed expressions in Algorithm~\ref{alg:count_distinct_2}.

\vspace{-1mm}
\subsection{Why standard self-attention is NOT able to learn \textit{count distinct} expression well?}
\label{sec:explain_sa_fail}
\vspace{-1mm}

Before we explain the reason, let us recap the formulation of standard self-attention network.
Let $\boldsymbol{c} \in \mathbb{R}^{k \times \tau}$ be a sequence of \textit{event-level} embeddings, where $\tau$ refers to the sequence length and $k$ refers to the embedding size.
Define $\boldsymbol{w_q} \in \mathbb{R}^{d_s \times k}, \boldsymbol{w_k} \in \mathbb{R}^{d_s \times k}, \boldsymbol{w_v} \in \mathbb{R}^{d_s \times k}$ be the trainable projection parameters.
A masked softmax self-attention neural network learns \textit{sequence-level} embedding by
\begin{flalign}
\label{eq:attention}
  \begin{aligned}
Attention\Big(\boldsymbol{c}; \boldsymbol{w_q}, \boldsymbol{w_k}, \boldsymbol{w_v}\Big) = 
\underbrace{Pooling\Bigg(\text{softmax}\bigg(\mathlarger{I}\Big(\frac{(\boldsymbol{w_q}\boldsymbol{c})^{T} (\boldsymbol{w_k}\boldsymbol{c})}{\sqrt{d_{s}}}, \boldsymbol{mask} \Big) \bigg) \bigg(\boldsymbol{w_v}\boldsymbol{c} \bigg)^{T}\Bigg)}_{\text{one-stage aggregation, $e.g.$, average or sum pooling.}}
  \end{aligned}
\end{flalign}
where 
\begin{flalign}
\label{eq:mask}
  \begin{aligned}
I(\boldsymbol{x}^{T}, \boldsymbol{mask}) = 
    \begin{cases}
        \boldsymbol{x_i} & \text{if~} mask_i=1 \\
        \boldsymbol{-\infty} & \text{otherwise}
    \end{cases}
  \end{aligned}
\end{flalign}
and $\boldsymbol{mask}$ is a vector indicating whether $i$-th event is a valid event.\footnote{In mini-batch training, different samples within the same batch could have different sequence length. We use truncate and padding operation on each sequence to make them have the same length. The mask vector is used to indicate the valid positions.}

\textbf{Explanation.}
According to Algorithm~\ref{alg:count_distinct_2}, the conditional \textit{count distinct} expression needs to memorize the frequency for each entity, $e.g.$, city, under some conditions and then count the number of entities with positive frequency.
There are two-stage aggregations in the \textit{count distinct} expression.
According to equation~\eqref{eq:attention},  the self-attention~(SA) neural network contains an explicit aggregation stage and there is no explicit clue that it can learn two-stages aggregations well.
An intuitive approach to conduct two-stage aggregations is using two-layer stacked self-attention~(SSA) network, $i.e.$, the output of the first SA network is the input of the second SA network.
However, such an approach also fails to learn the \textit{count-distinct} expressions well.
We find this failure is reasonable after carefully studying Algorithm~\ref{alg:count_distinct_2}.
Specifically, the first symbolic expression in Algorithm~\ref{alg:count_distinct_2} needs to \textit{memorize} the state for each entity, then the second symbolic expression learns the exact value. 
In contrast, although the SSA network conducts two-stage aggregation, the first layer cannot exactly \textit{memorize} the state (statistics) for each entity.
Then it's interesting to ask another question:

\begin{center}
\vspace{-1mm}
  \framebox{
  \parbox{0.9\columnwidth}{ 
  {\bf Question 2:} {Can we design novel neural architecture to learn symbolic expressions of conditional count distinct well?}    
  }}
\vspace{-1mm}
\end{center}

\subsection{Designing conditional sequence network}
\label{sec:design_cond_seq}

\begin{figure}[h!]
\vspace{-2mm}
\centering
\includegraphics[width=0.8\textwidth]{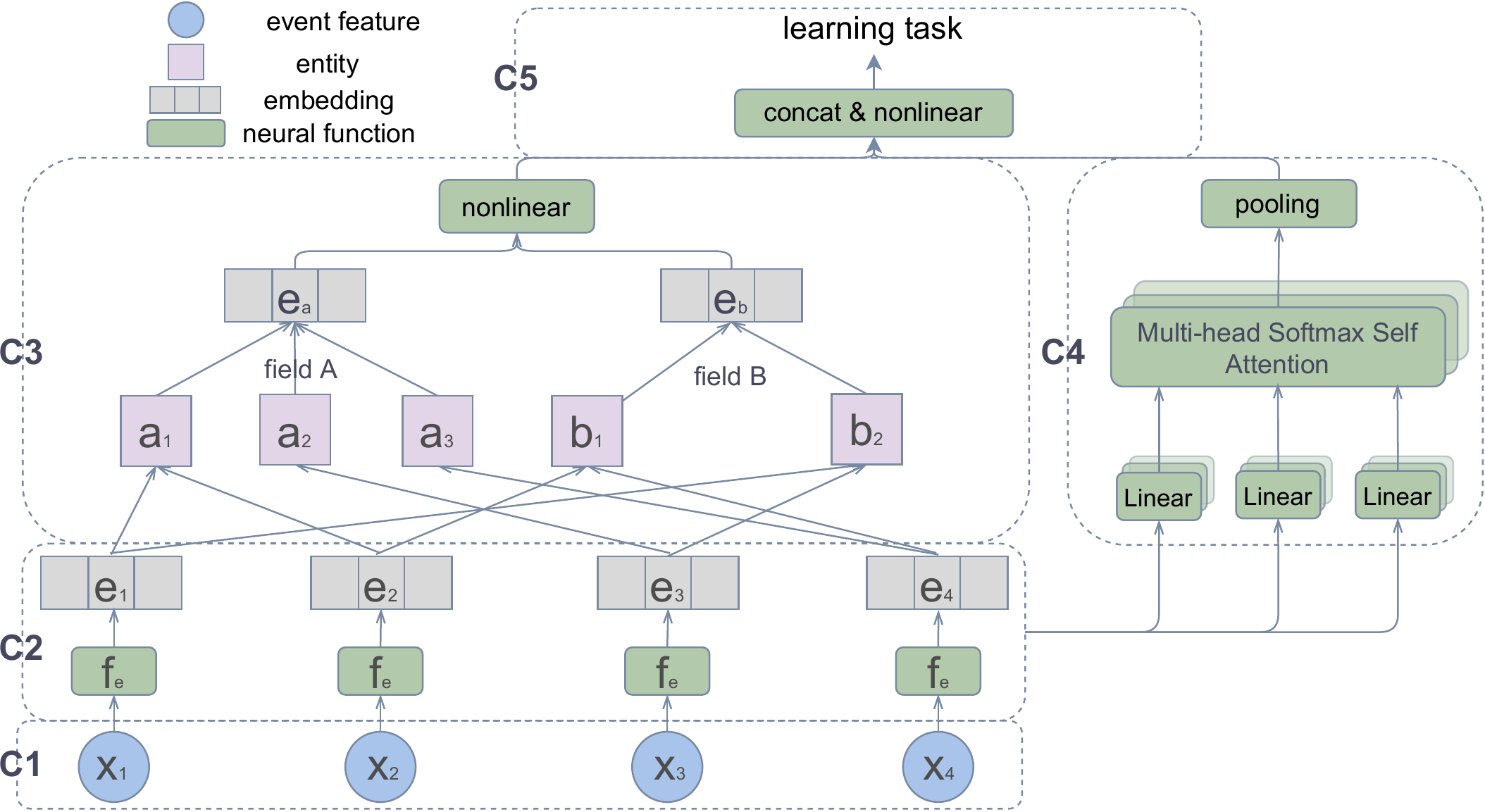}
\caption{\small The neural architecture of SHORING. 
C1: feature encoding for unstructured sequence data. 
C2: a provable high-order interaction network to learn \textit{event-level} embeddings.
C3 and C4 are our hybrid sequence networks used to learn \textit{sequence-level} embeddings.
C3: a novel symbolic-based conditional sequence network.
C4: multi-head self-attention network.
C5: task-specific network.
For example, in C3, there are two fields: A and B, where A has 3 entities and B has 2 entities, $e.g.$, filed A refers to city and 3 entities refer to New York, London and Shanghai. 
}
\label{fig:shoring}
\vspace{-4mm}
\end{figure}

\begin{algorithm}[H]
	\caption{Conditional Count Distinct (SQL)}
	\label{alg:count_distinct_1}
	\KwInput{sequence data saved in SQL table.}
	\KwOutput{z}
    z = \text{count(distinct if 1$\le t \le$5 then city else null end)}
\end{algorithm}

\begin{algorithm}[H]
	\caption{Conditional Count Distinct (symbolic)}
	\label{alg:count_distinct_2}
	\KwInput{a sequence of events}
	\KwOutput{z}
	\text{city\_list} = $\{\}$, \text{the initial value is 0};\\
	\text{//symbolic 1: conditional entities via group-by} \\
    \For{$\text{event in the sequence}$}{
        get the event time and the city;\\
        \text{city\_list[city]} += 1 if $1 \le t \le 5$ else 0.
    }
    \text{//symbolic 2: conditional fields via sum} \\
    z = sum([1 for freq in city\_list if freq>0])
\end{algorithm}

To answer question 2, as illustrated in figure~\ref{fig:shoring} (C3), we borrow the idea from symbolic learning~\citep{garcez2015neural} and design a novel neural network which approximates conditional statistics of entities as well as solves the two-stage aggregation problem.
Specifically, given the sequence data, we use a provable high-order interaction network to learn a sequence of \textit{event-level} embeddings $\boldsymbol{c} \in \mathbb{R}^{k \times \tau}$, which will be introduced in section~\ref{sec:prov_highorder}.
According to Theorem~\ref{thm:high_order}, it's expected that the neural network can learn the expert-defined conditions well, such as \textit{1 $\le t \le$ 5}.
We will test this assumption in the experiment.

\begin{algorithm}[H]
	\caption{Conditional Sequence Network}
	\label{alg:exact_count_distinct}
	\KwInput{event embedding: $\boldsymbol{c}$, assignment matrix: $\boldsymbol{p}$}
	\KwOutput{conditional embedding $\boldsymbol{z}$}
    \text{learning conditional entities} $\boldsymbol{e^{(e)}}=\sigma(\boldsymbol{p}\boldsymbol{c^{T}};\boldsymbol{w_p}, \boldsymbol{w_{p0}})$ \\
    \For{$\text{each field}~i$}{
    \text{learning} $\boldsymbol{e^{(f)}_i}=\sigma \Big(\sum_{\forall j, s.t., \boldsymbol{p}_{ij}=1} (\boldsymbol{e^{(e)}})_j;\boldsymbol{w_f}, \boldsymbol{w_{f0}} \Big)$; \\
    }
    $\boldsymbol{z} = \sigma \Big(\{\boldsymbol{e^{(f)}_i}\};\boldsymbol{w_z}, \boldsymbol{w_{z0}} \Big)$;

\end{algorithm}

The first symbolic expression in Algorithm~\ref{alg:count_distinct_2} conducts a \textit{group-by} operator and obtains frequency for each entity~\footnote{We define entity as a particular categorical value or id, such as city, gender, card type, or device id.}.
Define $\sigma$ as a nonlinear function.
Define $\boldsymbol{p} \in \mathbb{R}^{d_p \times \tau}$ as the \emph{assignment matrix}, where $\boldsymbol{p}_{ij}=1$, \text{if $j$-th event contains $i$-th entity}, otherwise $\boldsymbol{p}_{ij}=0$.
$d_p$ refers to the total number of entities within all the fields.
For example, in figure~\ref{fig:shoring}, $d_p=5$, $\tau=4$.
The term $\boldsymbol{p}\boldsymbol{c^{T}}$ could thus be considered as an approximation to the conditional \textit{entity-level} characteristics (\textit{c.f.} line 5 in Algorithm~\ref{alg:count_distinct_2}, $a_i$ and $b_j$ in C3 of figure~\ref{fig:shoring}).
Note that, in real-world applications, $\boldsymbol{p}$ is a sparse matrix and computing $\boldsymbol{p}\boldsymbol{c^{T}}$ is highly efficient.  
Then we conduct a sum pooling operator on conditional entities and obtain the conditional embeddings for fields (\textit{c.f.} line 8 in Algorithm~\ref{alg:count_distinct_2}, $e_a$ and $e_b$ in C3 of figure~\ref{fig:shoring}).
Finally, we use a nonlinear function to learn the conditional embedding $\boldsymbol{z}$ from the learned conditional fields $\{\boldsymbol{e^{(f)}_i}\}$.
We summarize the conditional sequence model in Algorithm~\ref{alg:exact_count_distinct}.

\vspace{-2mm}
\section{Learning High Order Interaction via Provable Reparameterization Trick}
\label{sec:prov_highorder}

In section~\ref{sec:symbolic_testing}, we introduce a helpful tool named symbolic testing and design a novel conditional sequence network to help us learn symbolic expressions of \textit{count distinct} from a sequence of \textit{event-level} embeddings.
We assume that the \textit{event-level} embeddings can cover the information of symbolic conditions we want to learn, such as conditions like $1\le  t \le 5$ in Algorithm~\ref{alg:count_distinct_1}.
Ideally, we expect the neural network can learn arbitrarily high-order feature interactions within limited time and memory resources.
In this section, we introduce a novel high-order interaction network based on an interesting provable reparameterization trick, all proofs will be deferred to appendix \ref{sec:appendixA}.

Define $\boldsymbol{a} \circ \boldsymbol{b}$ as the Hadamard product of vector $\boldsymbol{a}$ and $\boldsymbol{b}$, $i.e.$, for $\boldsymbol{a}, \boldsymbol{b}, \boldsymbol{c} \in \mathbb{R}^{k}$, $\boldsymbol{c} = \boldsymbol{a} \circ \boldsymbol{b}$, with the element given by $c_i=a_i*b_i$.
Define $<\boldsymbol{a}, \boldsymbol{b}>$ as the Frobenius inner product, which is the sum of the entries of the Hadamard product $\boldsymbol{a} \circ \boldsymbol{b}$, $i.e.$, $<\boldsymbol{a}, \boldsymbol{b}> = \sum^{k}_{i=1} a_i*b_i$.
For a set of vectors $ \{\boldsymbol{v_i}\}_{i=1}^n $, define $<\boldsymbol{v_1}, \ldots, \boldsymbol{v_n}> = \sum_{l=1}^k( \boldsymbol{v_1} \circ \ldots, \circ \boldsymbol{v_n})_l$ as the coordinate wise sum of Hadamard product.
Denote $\boldsymbol{x} \in \mathbb{R}^{d}$ as the feature vector, where $x_i$ refers to the $i$-th value in $\boldsymbol{x}$. 
Let $g(x) \in \mathbb{R}$ refer to a parameterized function used to learn feature interactions up to $ d $-th order.
In general, we formulate $g(\boldsymbol{x})$ by
\begin{flalign}
\label{eq:general_high_order}
\vspace{-2mm}
  \begin{aligned}
g(\boldsymbol{x}) := \underbrace{<\boldsymbol{w^{(1)}}, \boldsymbol{x}>}_{linear (1st~order)} 
     + \underbrace{\sum_{i_1<i_2}<\boldsymbol{w^{(2)}_{i_1}}, \boldsymbol{w^{(2)}_{i_2}}>x_{i_1} x_{i_2}}_{2nd~order} 
     + \cdots 
     + \underbrace{\sum_{i_1<\cdots<i_d}<\boldsymbol{w^{(d)}_{i_1}}, \boldsymbol{w^{(d)}_{i_2}}, \cdots, \boldsymbol{w^{(d)}_{i_d}}>x_{i_1} x_{i_2} \cdots x_{i_d}}_{d-th~order}
  \end{aligned}
\end{flalign}
where the parameters $\boldsymbol{w^{(1)}} \in \mathbb{R}^{d}, \boldsymbol{w^{(j)}_i} \in \mathbb{R}^{k}$.
Let $w^{(j)}_{il}$ denote the $l$-th element in $\boldsymbol{w^{(j)}_i}$, $i \in [1, d], j \in [2, d], l \in [1, k]$.

Eq.~\ref{eq:general_high_order} provides a general formulation for learning $d$-th order feature interactions. When $ d=2 $, Eq.\ref{eq:general_high_order} reduces to the formulation of factorization machines (FM)~\citep{rendle2010factorization} (see Lemma~\ref{lem:2order}).

\begin{restatable}{lem}{lem2order}
\label{lem:2order}
(Linear-Time 2nd Order Interaction by~\citet{rendle2010factorization}).
The 2nd order feature interaction can be computed in linear time, that is,  
\begin{flalign}
  \begin{aligned}
\sum_{i_1<i_2}<\boldsymbol{w^{(2)}_{i_1}}, \boldsymbol{w^{(2)}_{i_2}}>x_{i_1} x_{i_2}= \sum^{k}_{l=1} \underbrace{\frac{1}{2} \Big[ \Big( \sum^{d}_{i=1} w^{(2)}_{il} x_i \Big)^2 - \sum^{d}_{i=1}\Big(w^{(2)}_{il} x_i \Big)^2 \Big]}_\text{2nd-order embedding} = \sum^{k}_{l=1} \Big(h_{2nd} (\boldsymbol{x}) \Big)_l
  \end{aligned}
\end{flalign}

where $\big(\boldsymbol{x} \big)_l$ refers to $l$-th element in $\boldsymbol{x}$ and $h_{2nd}(\boldsymbol{x})$ is the embedding function for the 2nd-order interaction

\begin{flalign}
  \begin{aligned}
h_{2nd}(\boldsymbol{x}):= \frac{1}{2} \Big[ \Big( \sum^{d}_{i=1} w^{(2)}_{il} x_i \Big)^2 - \sum^{d}_{i=1}\Big(w^{(2)}_{il} x_i \Big)^2 \Big].
  \end{aligned}
\end{flalign}

\end{restatable}

Despite the potentially rich information brought by higher-order interactions, the prohibitive computational overhead makes high-order FMs impractical. Recently, inspired by logarithmic neural network (LNN)~\cite{hines1996logarithmic}, \citet{cheng2020adaptive} use a logarithmic transformation layer to learn high-order interaction and achieve promising results. Here we provide an interpretation of the logarithmic transformation layer as a simple consequence of the \emph{reparameterization trick}. To make a friendly derivation, we omit the superscript $(d)$ in $\boldsymbol{w^{(d)}_{i}}$ and $w^{(d)}_{ij}$ to simplify the notations.

\begin{restatable}{lem}{lemreptrick}
\label{lem:rep_trick2}
(Reparameterization Trick).
Define $x_i, x_j \in \mathbb{R^{+}}, w_{i} \in (\mathbb{R^{+}})^k$. 
For \textit{d}-th order interaction, there exists $\boldsymbol{v_{i_j}}$, $s.t.$, 
$$<\boldsymbol{w_{i_1}}, \boldsymbol{w_{i_2}}, \cdots, \boldsymbol{w_{i_d}}>x_{i_1} x_{i_2} \cdots x_{i_d} = \sum^{k}_{l=1}(x^{\boldsymbol{v_{i_1}}}_{i_1} \circ x^{\boldsymbol{v_{i_2}}}_{i_2} \circ \cdots \circ x^{\boldsymbol{v_{i_d}}}_{i_d})_l$$ 
\end{restatable}

According to lemma ~\ref{lem:rep_trick2}, we reparameterize $\boldsymbol{w^{(j)}_i}$ in Eq.~\eqref{eq:general_high_order} by $\boldsymbol{v^{(j)}_i}$ and transform the problem of learning high-order interaction to the problem of learning exponential embedding~\footnote{The exponential embedding in this paper is different from the exponential family embeddings~\citep{rudolph2016exponential}, which refers to a class of methods that extends the idea of word embeddings to other types of
high-dimensional data.}.

\begin{restatable}{thm}{thmphighorder}
\label{thm:high_order}
(Arbitrary High-Order Interaction) Assume all the elements in $\boldsymbol{x}$ are positive, with the reparameterization trick in Lemma~\ref{lem:rep_trick2}, the Eq.~\ref{eq:general_high_order} can be written by 
\begin{flalign}
\label{eq:re_general_high_order}
  \begin{aligned}
g(\boldsymbol{x}) &:= \underbrace{<\boldsymbol{w^{(1)}}, \boldsymbol{x}>}_{linear (1st~order)} 
     + \underbrace{\sum_{i_1<i_2}<\boldsymbol{w^{(2)}_{i_1}}, \boldsymbol{w^{(2)}_{i_2}}>x_{i_1} x_{i_2}}_{2nd~order} 
     + \cdots 
     + \underbrace{\sum_{i_1<\cdots<i_d}<\boldsymbol{w^{(d)}_{i_1}}, \boldsymbol{w^{(d)}_{i_2}}, \cdots, \boldsymbol{w^{(d)}_{i_d}}>x_{i_1} x_{i_2} \cdots x_{i_d}}_{d-th~order} \\
&= \boldsymbol{w^{(1)}} \boldsymbol{x} 
     + \sum^{k}_{l=1}\Bigg(\underbrace{\sum_{o=1}^{|O|}\Big(exp\big(\boldsymbol{u_{o}}\ln{\boldsymbol{x}}\big)\Big)}_{high-order~embedding~summation} \Bigg)_l \\
&= \boldsymbol{w^{(1)}} \boldsymbol{x} 
     + \sum^{k}_{l=1}\Bigg(\sum_{o=1}^{|O|} \big(h_o(\boldsymbol{x}; \boldsymbol{u_{o}} \big) \Bigg)_l
  \end{aligned}
\end{flalign}
where $O=\big\{i_1<i_2\} \cup \{i_1<i_2<i_3\} \cup \cdots \cup \{i_1<\cdots<i_d \big\}$ refers to the set of cases of all feature interactions, $\boldsymbol{x} \in \mathbb{R}^{d}$, $\{\boldsymbol{u_{o}} \in \mathbb{R}^{k \times d}| o \in [1, |O|]\}$ refers to the set of parameters, $h_o(\boldsymbol{x}; \boldsymbol{u_{o}}):= exp\big(\boldsymbol{u_{o}}\ln{\boldsymbol{x}}\big)$ is the $o$-th embedding function for the high order interaction.

\end{restatable}

\textbf{Remark.}
Define $h_o(\boldsymbol{x};\boldsymbol{u_{o}})=exp\big(\boldsymbol{u_{o}}\ln{\boldsymbol{x}}\big) \in \mathbb{R}^{k}$, which can be considered as the learned exponential embedding of $\boldsymbol{x}$ with parameter $\boldsymbol{u_{o}}$ for $o$-th interaction case. Theorem~\ref{thm:high_order} transforms the high-order interaction problem to a very simple problem of learning summation over $\{h_o(\boldsymbol{x}; \boldsymbol{u_{o}}) | o \in [1, |O|] \}$, alleviating the heavy computational cost caused by explicitly calculating the high-order interactions, at the same time being implementation-friendly. Moreover, it also shed lights on the design of a more efficient neural architecture termed SHORING that we believe to be beneficial for sequence learning tasks.

\vspace{-1mm}
\section{SHORING: Multiple Sequences Model}
\label{sec:general_multseqmodel}
\vspace{-1mm}

Now it's ready to introduce our multiple sequences model: multiple Sequences High-ORder Interaction Network with conditional Group embedding (SHORING).
Define $\sigma(\boldsymbol{x})$ as a function to learn the nonlinear transformation of $\boldsymbol{x}$. Specifically, $\sigma(\boldsymbol{x})$ is a shallow neural network, $i.e.$, $\sigma(\boldsymbol{x}; \boldsymbol{w}, \boldsymbol{w_{0}}) = \textit{relu}(\boldsymbol{w}\boldsymbol{x} + \boldsymbol{w_{0}})$, where $relu(x)=max(x, 0)$.

\vspace{-1mm}
\subsection{Event-level representation}
\vspace{-1mm}

Let $\boldsymbol{x_{ij}} \in \mathbb{R}^{d_{e}}$ refer to the $j$-th event feature in $i$-th sequence as illustrated by figure~\ref{fig:multiple_sequence}.
We keep a \emph{vector} $ \textit{mask} \in \{0, 1\}^{\tau} $ for each sequence, where $mask[i]=0$ refers to a padding event $i$.  
We transformer the feature by $\boldsymbol{\tilde{x}_{ij}}=\sigma(\boldsymbol{x_{ij}}; \boldsymbol{w_x}, \boldsymbol{w_{x0}}) + \epsilon$, where $\epsilon$ is a small positive scalar.
Hereafter we will drop the sequence index $i$ and $j$ for clarity of presentation.
After the nonlinear transformation, we have $\tilde{\boldsymbol{x}} \in \mathbb{R}_{+}^{d_{e} \times \tau \times m}$, $i.e.$, 
all the elements in $\tilde{\boldsymbol{x}}$ are positive. The assumption of Theorem~\ref{thm:high_order} holds. Define $\{\boldsymbol{u_{o}} \in \mathbb{R}^{k \times d_e}| 1 \le o \le |O_e| \}$, where $O_e=\{i_1<i_2\} \cup \{i_1<i_2<i_3\} \cup \cdots \cup \{i_1<\cdots<i_{d_e}\}$.

Typically, explicit linear transformation and \textit{2nd} order feature interactions have achieved much success in real-world applications.
According to Lemma~\ref{lem:2order}, we compute \textit{2nd}-order feature interaction within the linear time and memory. 
Our event network is defined by
\begin{flalign}
\label{eq:high_order_emd}
  \begin{aligned}
f_e(\boldsymbol{x};\theta) := \sigma\Bigg(
\boldsymbol{w^{(1)}}\boldsymbol{\tilde{x}} + h_{2nd}(\boldsymbol{\tilde{x}})
+ \sum^{N}_{o=3} h_o(\boldsymbol{u_{o}}, \boldsymbol{\tilde{x}}) ; \boldsymbol{w_e}, \boldsymbol{w_{e0}} \Bigg).
  \end{aligned}
\end{flalign}
where $\theta=\big\{ \boldsymbol{w_{x}^{(1)}}, \boldsymbol{w_{x}^{(0)}} \boldsymbol{w^{(1)}}, \boldsymbol{w^{(2)}}, \boldsymbol{w_e}, \boldsymbol{w_{e0}, \boldsymbol{u_{o}}} | o \in [1, N] \big\}$ is the set of trainable parameters, $\boldsymbol{w^{(1)}}, \boldsymbol{w^{(2)}} \in \mathbb{R}^{k \times d_{e}}$, $w^{(2)}_{il}$ refers to the the element of $\boldsymbol{w^{(2)}}$ at $i$-th row and $l$-th column.

\textbf{Remark.}
In practice, we find that specifying a small $N$ in Eq.~\eqref{eq:high_order_emd} is good enough for the real-world applications. 
Typically,  we use the parameter-sharing technique to reduce the number of parameters, $i.e.$, using the same parameters for all the events within the same sequence.

\vspace{-1mm}
\subsection{Sequence-level representation}
\vspace{-1mm}

Eq.\ref{eq:high_order_emd} provides us all event-level embeddings within the multiple sequences.
We use the proposed conditional sequence network in Algorithm~\ref{alg:exact_count_distinct} and multi-head self-attention network ($H$ heads) to learn sequence-level embedding by 
\begin{flalign}
\vspace{-1mm}
\label{eq:hybird_seq}
  \begin{aligned}
f_s(s;~\beta) := \Big[\sigma \big(\boldsymbol{z};\boldsymbol{w_z}, \boldsymbol{w_{z0}} \big), 
\frac{1}{H} \sum^{H}_{i=1} Attention_{i}\Big(\boldsymbol{c}; \boldsymbol{w_{iq}}, \boldsymbol{w_{ik}}, \boldsymbol{w_{iv}} \Big) \Big].
  \end{aligned}
\vspace{-1mm}
\end{flalign}

\vspace{-3mm}
\subsection{Learning Task}
\vspace{-1mm}

In general, we employ a nonlinear neural network, such as multilayer perceptrons, to learn the specific task from the concatenate sequence embeddings based on the defined loss function. 
Specifically, the mean-squared loss is used to learn regression tasks and Kullback-Leibler (KL) divergence is used to learn multi-class classification tasks.

\vspace{-2mm}
\section{Experiment}
\label{sec:experiment}
\vspace{-2mm}

We compare the proposed method against nine state-of-the-art baselines on four synthetic datasets, a public dataset and two industrial datasets.
The results show that our method empirically consistently outperforms all baselines.
The ablation studies show that both the proposed high-order interaction network and the conditional sequence network are very helpful.
We have deployed the industrial models on one of the largest online trade systems.

\vspace{-2mm}
\subsection{Datasets}
\label{sec:dataset}
\vspace{-2mm}

\begin{table}[]
\caption{\small Statistics of different datasets. Task: R refers to regression task, C refers to classification task.
}
\small
\centering
\begin{threeparttable}
\begin{tabular}{lcccccc}
\toprule
    dataset     & task  & events & samples & \#fields & \# total unique categories & \# total dense features \\ \midrule
Synthetic-1 & R  & 1,544,955 & 10000 & 13 & 70 & 4 \\ 
Synthetic-2 & R  & 1,544,955 & 10000 & 13 & 80 & 4 \\ 
Synthetic-3 & R  & 1,544,955 & 10000 & 13 & 100 & 4 \\ 
Synthetic-4 & R  & 1,544,955 & 10000 & 13 & 160 & 4 \\ 
\hline
Taobao & C & 100,150,807 & 987,994 & 3 & 4171401 & 0 \\
\hline
Financial-1 & C  & 78 million & 7,822,536 & 47 & 9,293,711 & 3 \\ 
Financial-2 & C  & 1.3 billion & 952,271  & 28 & 2548 & 4 \\ 
\bottomrule
\end{tabular}
 \end{threeparttable}
\label{tb:dataset}
\vspace{-4mm}
\end{table}

\textbf{Synthetic Datasets.}
We assume each user's behavior sequence is independent and each event contains 9 categorical attributes and 4 numerical attributes. The distribution of event data follows the prior distribution learned from the real-world behavior sequences.
We randomly generate four sets of sequences: \textit{Synthetic-1}, \textit{Synthetic-2}, \textit{Synthetic-3}, and \textit{Synthetic-4}, with the first categorical field contains 10, 20, 50 and 100 different unique values and other fields being the same. Each sequence length is randomly chosen to be between 10 and 300. The details of these synthetic datasets are summarized in table~\ref{tb:dataset}.
To conduct symbolic testing as illustrated in table~\ref{tb:symbolic_testing}, we make 13 regression tasks, each of which corresponds to applying a specific operator to a derived set of the input sequence. The derivation rule is briefed as follows: For all but the \textit{count distinct} operator, we select the first and third numerical attributes, and randomly generates a set of interactions between them, and apply computes the sequence-level summary using the specified operator. 
For the count distinct operator, we apply it to the first categorical feature to produce the regression target.

\textbf{Real-world datasets.}We use one public dataset and two industrial datasets to validate the performance of the proposed method.\textbf{Taobao} is a public dataset containing a collection of users' behaviors, $i.e.$, click, purchase and $etc$, from Taobao's recommendation system. The learning task is to predict whether users will click $t$-th product when given the past $t-1$ behaviors.
We filter the samples whose length is short than 20 and truncate sequence at length 200. \textbf{Financial-1} is an industrial dataset collected from the online transaction system of one of the world-leading cross-border e-commerce platforms. Samples are constructed from user payment event data when they purchase goods (the resulting sequence that is referred to as \emph{behavior sequences}). 
Under a maximum sequence length of $ 40 $, the task is to detect whether the current payment is a fraud. The fraud labels are collected from the chargeback reports from card issuer banks.
\textbf{Financial-2} is an industrial dataset collected from an online transaction system.
The dataset is organized like \textbf{Financial-1}, whose maximum sequence length is $ 300 $. The task is to predict whether the given event is a fraud based on four sequences of behaviors within the last 7 days. The total number of events for this scenario is 1.3 billion. Detailed characteristics of the datasets are listed in table \ref{tb:dataset}.

\vspace{-2mm}
\subsection{Experimental setups}
\vspace{-2mm}

\textbf{Evaluation Metric}.
For symbolic testing on synthetic datasets, we report seven goodness-of-fit metrics that measures how well the specific neural model fits the underlying data: 
\textit{loss} (mean-squared loss), 
\textit{std\_r=$\frac{|\text{std} - \hat{\text{std}}|}{\text{std}}$} ($\text{std}$ and $\hat{\text{std}}$ are the standard deviation of predictions and targets), \textit{ptb@1\%}(the number of samples whose perturbation is larger than 1\%), 
\textit{ptb\_r@1\%}(the average perturbation of samples whose perturbation is larger than 0.01), 
$R^{2}$ (R-squared),
\textit{pearson} correlation between predictions and targets.
In general, lower \textit{loss}, \textit{std\_r}, \textit{ptb@1\%}, \textit{ptb\_r@1\%} and higher $R^{2}$, \textit{pearson} will be better.
We also employ the kernel two-sample hypothesis test~\citep{gretton2012kernel}, which is a nonparametric method to compare samples from two probability distributions. We report the \textit{p-value} of the test.
Note that, the null hypothesis of kernel two-sample test is both samples stem from the same distribution. We treat tests with p-value smaller than $ 0.05 $ as a statistical evidence of failing to fit the target.
For the classification tasks on real-world datasets, we use Kullback-Leibler (KL) divergence as the optimization loss.
We report KL loss, AUC score, and recall under $ 99\% $ precision.

\begin{table}[t!]
\caption{\small Symbolic Testing on synthetic datasets. The standard softmax self-attention network can NOT learn conditional symbolic expression well.
The first 10 testings are conducted on \textit{Synthetic-1}, and the others are on \textit{Synthetic-2,3,4} respectively.
} 
\label{tb:symbolic_testing}
\small
\setlength{\tabcolsep}{2.6pt}
\begin{center}
\begin{threeparttable}
\begin{tabular}{c|ccccccc}
\toprule
Symbolic Exp. & loss$\downarrow$  & std\_r$\downarrow$ & ptb@1\%$\downarrow$ & ptb\_r@1\%$\downarrow$ & $R^{2}$$\uparrow$ & pearson$\uparrow$  & p-value$\uparrow$ \\
\hline
sum & 0.0035 & 0.0025 & 0 & 0 & 0.9999 & 0.9999  & 1.00  \\
count & 0.0019 & 0.0007 & 0 & 0 & 1.0000 & 1.0000  & 1.00  \\
average & 0.0003 & 0.0061 & 0 & 0 & 0.9997  & 0.9998 & 1.00  \\
decay sum & 0.0001 & 0.0012 & 0 & 0 & 0.9997 & 0.9999  & 1.00  \\
decay count & 0.0034 & 0.0019 & 0 & 0 & 0.9997 & 0.9999  & 1.00  \\
decay average & 0.0003 & 0.0102 & 0 & 0 & 0.9989 & 0.9995  & 1.00  \\
sum/sum & 0.0027 & 0.0033 & 0 & 0 & 0.9992  & 0.9996 & 0.99  \\
count/count & 0.0009 & 0.0027 & 0 & 0 & 0.9994 & 0.9997  & 1.00  \\
sum/average & 0.0085 & 0.0021 & 0 & 0 & 0.9992 & 0.9996  & 0.99  \\
\hline
distinct(10) & 0.2504 & 0.2751 & 0.9420 & 0.1140 & 0.4479 & 0.6692  & 0.01  \\
distinct(20) & 0.2560 & 0.2569 & 0.8850 & 0.1130 & 0.5474 & 0.7398  & 0.01  \\
distinct(50) & 0.2563 & 0.2383 & 0.8930 & 0.1050 & 0.5252 & 0.7247  & 0.01  \\
distinct(100) & 0.2566 & 0.3139 & 0.8530 & 0.1060 & 0.5033 & 0.7094  & 0.01  \\
\bottomrule
\end{tabular}
\vspace{-4mm}

\end{threeparttable}
\end{center}
\vspace{-4mm}
\end{table}

\textbf{Baselines}.
We compare the proposed SHORING against 9 state-of-the-art end-to-end feature interaction models, including:
(1) W\&D~\citep{cheng2016wide}; (2) DeepFM~\citep{guo2017deepfm}; (3) AFM~\citep{xiao2017attentional};
(4) Product neural network~(PNN)~\citep{qu2016product};
(5) HOFM~\citep{blondel2016higher};
(6) DCN~\citep{wang2017deep};
(7) AFN~\citep{cheng2020adaptive};
(8) AutoInt~\citep{song2019autoint};
(9) DIN~\citep{zhou2018deep}.
These baselines achieved state-of-the-art performance in many end-to-end learning tasks.
We introduce the details about each baseline in appendix~\ref{sec:baseline}.

\vspace{-2mm}
\subsection{Symbolic Testing on Synthetic Datasets}
\label{sec:symbolic}
\vspace{-1mm}

\textbf{Testing standard softmax self-attention neural network}
We conduct a set of symbolic testing for standard Self-Attention (SA) network on the simple synthetic datasets.
Table~\ref{tb:symbolic_testing} reports the results (see SA in Table~\ref{tb:symbolic_testing}).
Although it performs well for several standard symbolic expressions, such as \textit{sum, count, average} and their time-decay and ratio versions, it fails to learn the conditional \textit{count-distinct} expressions, including \textit{distinct(10)}, \textit{distinct(20)}, \textit{distinct(50)} and \textit{distinct(100)}.
Specifically, the kernel two-sample hypothesis testing~\citep{gretton2012kernel} rejects the null hypothesis that both samples stem from the same distribution (\textit{p-value}=$0.01$).
We also conduct symbolic testing for two-layer Stacked Self-Attention (SSA), however, it also fails (see SSA in Table~\ref{tb:symbolic_testing}).
We provide the explanation in section~\ref{sec:explain_sa_fail}.

\textbf{Testing our conditional sequence model}
We evaluate the proposed conditional sequence network on the same datasets as present the results in table \ref{tb:symbolic_testing_improve}.
The results show our method learns these 13 symbolic expressions well (much better than both SA and SSA). 
Such success might suggest symbolic testing is a good tool to evaluate neural networks and the idea of symbolic learning can help us design more powerful neural architecture.

\begin{table}[t!]
\vspace{-2mm}
\caption{\small Comparison of regression tasks via Symbolic Testing: count distinct expressions on \textit{Synthetic-{1,2,3,4}}. SA: standard self-attention, SSA: stacked self-attention, our method SHORING performs best.
} 
\label{tb:symbolic_testing_improve}
\small
\setlength{\tabcolsep}{3.5pt}
\begin{center}
\begin{threeparttable}
\begin{tabular}{c|ccccccc}
\toprule
Model(\#unique)~\tnote{a} & loss$\downarrow$  & std\_r$\downarrow$ & ptb@1\%$\downarrow$ & ptb\_r@1\%$\downarrow$ & $R^{2}$$\uparrow$ & pearson$\uparrow$  & p-value$\uparrow$ \\
\hline
SA(10) & 0.2504 & 0.2751 & 0.9420 & 0.1140 & 0.4479 & 0.6692  & 0.01  \\
SSA(10) & 0.2468 & 0.3184 & 0.9180 & 0.1160 & 0.4585 & 0.6771  & 0.01  \\
SHORING(10) & \bf0.0003 & \bf0.0101 & \bf0 & \bf0 & \bf0.9992 & \bf0.9996  & \bf1.00  \\
\hline
SA(20) & 0.2560 & 0.2569 & 0.8850 & 0.1130 & 0.5474 & 0.7398  & 0.01  \\
SSA(20) & 0.2541 & 0.2257 & 0.9110 & 0.1110 & 0.5598 & 0.7482  & 0.01  \\
SHORING(20) & \bf0.0006 & \bf0.0009 & \bf0 & \bf0 & \bf0.9992 & \bf0.9996  & \bf0.98  \\
\hline
SA(50) & 0.2563 & 0.2383 & 0.8930 & 0.1050 & 0.5252 & 0.7247  & 0.01  \\
SSA(50) & 0.2533 & 0.2947 & 0.9230 & 0.1040 & 0.5344 & 0.7310  & 0.01  \\
SHORING(50) & \bf0.0013 & \bf0.0077 & \bf0 & \bf0 & \bf0.9978 & \bf0.9989  & \bf0.96  \\
\hline
SA(100) & 0.2566 & 0.3139 & 0.8530 & 0.1060 & 0.5033 & 0.7094  & 0.01  \\
SSA(100) & 0.2542 & 0.3361 & 0.8700 & 0.1050 & 0.5066 & 0.7118  & 0.01  \\
SHORING(100) & \bf0.0005 & \bf0.0311 & \bf0 & \bf0 & \bf0.9953 & \bf0.9977  & \bf0.97  \\
\bottomrule
\end{tabular}

\begin{tablenotes}
\item[a] \scriptsize{\textit{\#unique}=$\{10, 20, 50, 100\}$ are the number of unique values for the first field on \textit{Synthetic-{1,2,3,4}}. }
\end{tablenotes}

\end{threeparttable}
\end{center}
\vspace{-6mm}
\end{table}

\vspace{-2mm}
\subsection{Performance on real-world datasets}
\vspace{-1mm}

Table~\ref{tb:public_data} reports the results for different methods on real-world datasets. Two variants of our proposed model are evaluated, namely SHORIN (corresponds to a reduced model \emph{without} the conditional sequence module) and SHORING which is our full model. Across all the three datasets, our full model SHORING exhibits consistently better performance under AUC and recall. The superiority is particularly evident under the recall metric, in that we achieve $ 19.2\% $ relative improvement over the strongest baseline. Moreover, the reduced model SHORIN performs competitively across all the tasks, on Taobao and Financial-2 it dominates all the baselines, thereby showing the effectiveness of the proposed higher-order interaction module. Finally we observe that the incorporation of the conditional sequence model yields solid improvements. On the Taobao dataset, the conditional sequence model results in a $ 4.2\% $ relative gain in the recall. 

\begin{table}[t!]
\caption{\small Comparison on public and industrial datasets with statistical significance testing.
We run all experiments with three random seeds and report the standard deviation for each metric on all datasets.
SHORING is the complete version of our method and SHORIN without the conditional sequence network in section~\ref{sec:design_cond_seq} is used to conduct ablation studies.
} 
\label{tb:public_data}
\small
\setlength{\tabcolsep}{1.6pt}
\begin{center}
\begin{tabular}{c|ccc|ccc|ccc}
\toprule
 & \multicolumn{3}{|c}{Financial-1} & \multicolumn{3}{|c}{Financial-2} & \multicolumn{3}{|c}{Taobao} \\
\hline 
Model & auc(\%)$\uparrow$  & loss(1e-3)$\downarrow$ & recall(\%)$\uparrow$ & auc(\%)$\uparrow$  & loss(1e-3)$\downarrow$ & recall(\%)$\uparrow$ & auc(\%)$\uparrow$  & loss(1e-2)$\downarrow$ & recall(\%)$\uparrow$  \\
\hline
W\&D & 93.1$\pm$0.2 & 11.5$\pm$0.0 & 67.8$\pm$0.3 & 96.6$\pm$1.1  & 26.0$\pm$3.7 & 62.5$\pm$3.1 & 93.5$\pm$0.3 & 16.2$\pm$0.4 & 39.1$\pm$0.8    \\
DeepFM & 93.3$\pm$0.1 & 13.1$\pm$0.1 & 65.9$\pm$0.2 & 96.3$\pm$0.7  & 41.8$\pm$5.4 & 66.1$\pm$3.4 & 93.4$\pm$0.2 & 16.3$\pm$0.5 & 38.0$\pm$0.7  \\
AFM & 92.0$\pm$0.3 & 12.2$\pm$0.2 & 62.2$\pm$0.3 & 96.6$\pm$0.5 & 26.9$\pm$3.0 & 60.2$\pm$5.7 & 92.9$\pm$0.4 & 16.8$\pm$0.4 & 35.4$\pm$0.3  \\
PNN & 93.2$\pm$0.2 & 11.2$\pm$0.1 & 67.2$\pm$0.2 & 96.2$\pm$0.4  & 32.9$\pm$2.9 & 57.6$\pm$3.5 & 94.0$\pm$0.2 & 15.5$\pm$0.3 & 38.7$\pm$0.8 \\
HOFM & 92.4$\pm$0.4 & 11.7$\pm$0.1 & 63.8$\pm$0.2 & 95.3$\pm$0.9  & 23.2$\pm$4.2 & 56.8$\pm$5.2 & 93.2$\pm$0.3 & 16.5$\pm$0.5 & 37.1$\pm$0.4  \\
DCN  & 94.1$\pm$0.1 & \bf10.3$\pm$0.1 & 69.2$\pm$0.4 & 96.7$\pm$0.5  & 25.7$\pm$3.0 & 68.9$\pm$7.0 & 93.9$\pm$0.2 & 15.6$\pm$0.2 & 39.1$\pm$0.6 \\
AFN & 92.8$\pm$0.4 & 15.6$\pm$0.3 & 62.3$\pm$0.2 & 96.5$\pm$0.8  & 28.0$\pm$5.1 & 65.7$\pm$2.8 & 93.0$\pm$0.3 & 16.9$\pm$0.3 & 36.3$\pm$0.4   \\
AutoInt & 91.7$\pm$0.2 & 16.4$\pm$0.3 & 57.8$\pm$0.5 & 95.9$\pm$0.4  & 24.6$\pm$2.2 & 60.0$\pm$4.1 & 93.4$\pm$0.4 & 16.5$\pm$0.3 & 36.7$\pm$0.3 \\
DIN & 93.4$\pm$0.2 & 10.4$\pm$0.1 & 69.3$\pm$0.2 & 96.5$\pm$0.8 & 22.3$\pm$2.8 & 67.2$\pm$2.5 & 94.0$\pm$0.2 & 15.5$\pm$0.3 & 39.5$\pm$0.4  \\

\hline
{SHORIN} & 93.9$\pm$0.1 & 11.5$\pm$0.2 & 68.4$\pm$0.1 & 98.9$\pm$0.4  & 9.7$\pm$0.7 & 81.5$\pm$4.3 & 94.4$\pm$0.3 & 15.1$\pm$0.2 & 42.7$\pm$0.3  \\
{SHORING} & \bf94.4$\pm$0.2 &  11.1$\pm$0.1 & \bf70.5$\pm$0.2 & \bf99.0$\pm$0.6  & \bf9.2$\pm$0.8 & \bf82.1$\pm$3.3 & \textbf{94.9$\pm$0.2} & \textbf{14.5$\pm$0.2} & \textbf{44.5$\pm$0.4}  \\
\bottomrule
\end{tabular}
\end{center}
\vspace{-4mm}
\end{table}
\vspace{-3mm}

\vspace{-1mm}
\section{Discussion, Conclusion and Future Work}
\vspace{-2mm}

\textbf{Discussion of differences among representation learning, neural architecture search and neural symbolic representation}.
In the field of machine learning and deep learning, there are many articles addressing representation learning, the goal of which is to learn effective feature representations from data via learning schemes, such as supervised learning~\cite{goodfellow2016deep}, self-supervised learning~\citep{vaswani2017attention}, and reinforcement learning~\cite{sutton1998introduction}.
The goal of neural architecture search (NAS) is to design (search for) effective combinations of neural units from the specific search space based on carefully designed search strategy and performance estimation method.
Although NAS has achieved many successes, its searching space depends on building blocks from a set of given neural units, which are typically neural operators selected from standard neural architectures, such as multilayer perceptron, Long short-term memory (LSTM)~\citep{1997Hochreiterlstm}, self-attention and transformer~\citep{vaswani2017attention}, $etc$.
Our neural symbolic representation (NSR) is more likely to provide additional neural units with new designs inspired by the idea of symbolic learning which could provide NAS with building blocks that are more effective to specific applications.
It's expected that with the the help of NAS, we can make more powerful neural architectures and achieve much better performance.

\textbf{Conclusion}.
In this paper, based on the proposed symbolic testing, we argue that standard self-attention neural network is not capable of learning certain conditional symbolic expressions well.
Inspired by symbolic learning, we proposed a novel multiple sequences model named SHORING based on the provable high-order interaction network and conditional sequence network, which has the representation power to effectively learn various symbolic expressions through sequence data.

\textbf{Future Work}.
In the future, under the same general frame work we will design long sequence model to capture user's lifelong behavior and expand tests to cover more types and combinations of symbolic expression.
Also, in this paper, we only talk about neural symbolic learning on multiple sequences data. Following the similar spirit of our neural symbolic representation, it's easy to extend our work to dynamic graph neural network~\citep{wang2021tcl}, which potentially could capture more complex structured and spatial interactions among different entities.

\clearpage
\bibliographystyle{style/icml2019}
\bibliography{reference}

\begin{thebibliography}{49}
\providecommand{\natexlab}[1]{#1}
\providecommand{\url}[1]{\texttt{#1}}
\expandafter\ifx\csname urlstyle\endcsname\relax
  \providecommand{\doi}[1]{doi: #1}\else
  \providecommand{\doi}{doi: \begingroup \urlstyle{rm}\Url}\fi

\bibitem[Blondel et~al.(2016{\natexlab{a}})Blondel, Fujino, Ueda, and
  Ishihata]{blondel2016higher}
Blondel, M., Fujino, A., Ueda, N., and Ishihata, M.
\newblock Higher-order factorization machines.
\newblock \emph{NeurIPS}, 29:\penalty0 3351--3359, 2016{\natexlab{a}}.

\bibitem[Blondel et~al.(2016{\natexlab{b}})Blondel, Ishihata, Fujino, and
  Ueda]{blondel2016polynomial}
Blondel, M., Ishihata, M., Fujino, A., and Ueda, N.
\newblock Polynomial networks and factorization machines: New insights and
  efficient training algorithms.
\newblock \emph{ICML}, 2016{\natexlab{b}}.

\bibitem[Chen \& Guestrin(2016)Chen and Guestrin]{chen2016xgboost}
Chen, T. and Guestrin, C.
\newblock Xgboost: A scalable tree boosting system.
\newblock In \emph{SIGKDD}, 2016.

\bibitem[Chen et~al.(2018)Chen, Li, Li, Jiang, and Song]{chen2018neural}
Chen, X., Li, S., Li, H., Jiang, S., and Song, L.
\newblock Neural model-based reinforcement learning for recommendation.
\newblock 2018.

\bibitem[Chen et~al.(2019)Chen, Li, Li, Jiang, Qi, and
  Song]{chen2019generative}
Chen, X., Li, S., Li, H., Jiang, S., Qi, Y., and Song, L.
\newblock Generative adversarial user model for reinforcement learning based
  recommendation system.
\newblock In \emph{ICML}, 2019.

\bibitem[Cheng et~al.(2016)Cheng, Koc, Harmsen, Shaked, Chandra, Aradhye,
  Anderson, Corrado, Chai, Ispir, et~al.]{cheng2016wide}
Cheng, H.-T., Koc, L., Harmsen, J., Shaked, T., Chandra, T., Aradhye, H.,
  Anderson, G., Corrado, G., Chai, W., Ispir, M., et~al.
\newblock Wide \& deep learning for recommender systems.
\newblock In \emph{Proceedings of the 1st workshop on deep learning for
  recommender systems}, pp.\  7--10, 2016.

\bibitem[Cheng et~al.(2020)Cheng, Shen, and Huang]{cheng2020adaptive}
Cheng, W., Shen, Y., and Huang, L.
\newblock Adaptive factorization network: Learning adaptive-order feature
  interactions.
\newblock In \emph{AAAI}, 2020.

\bibitem[Choromanski et~al.(2020)Choromanski, Likhosherstov, Dohan, Song, Gane,
  Sarlos, Hawkins, Davis, Mohiuddin, Kaiser, et~al.]{choromanski2020rethinking}
Choromanski, K., Likhosherstov, V., Dohan, D., Song, X., Gane, A., Sarlos, T.,
  Hawkins, P., Davis, J., Mohiuddin, A., Kaiser, L., et~al.
\newblock Rethinking attention with performers.
\newblock \emph{arXiv}, 2020.

\bibitem[Dai et~al.(2018)Dai, Li, Tian, Huang, Wang, Zhu, and
  Song]{dai2018adversarial}
Dai, H., Li, H., Tian, T., Huang, X., Wang, L., Zhu, J., and Song, L.
\newblock Adversarial attack on graph structured data.
\newblock In \emph{International conference on machine learning (ICML)}, pp.\
  1115--1124. PMLR, 2018.

\bibitem[Das et~al.(2017)Das, Sahoo, and Datta]{das2017survey}
Das, D., Sahoo, L., and Datta, S.
\newblock A survey on recommendation system.
\newblock \emph{International Journal of Computer Applications}, 160\penalty0
  (7), 2017.

\bibitem[Devlin et~al.(2018)Devlin, Chang, Lee, and Toutanova]{devlin2018bert}
Devlin, J., Chang, M.-W., Lee, K., and Toutanova, K.
\newblock Bert: Pre-training of deep bidirectional transformers for language
  understanding.
\newblock \emph{arXiv}, 2018.

\bibitem[Feng et~al.(2019)Feng, Lv, Shen, Wang, Sun, Zhu, and
  Yang]{feng2019deep}
Feng, Y., Lv, F., Shen, W., Wang, M., Sun, F., Zhu, Y., and Yang, K.
\newblock Deep session interest network for click-through rate prediction.
\newblock \emph{arXiv}, 2019.

\bibitem[Garcez et~al.(2015)Garcez, Besold, Raedt, Foldiak, Hitzler, Icard,
  Kuhnberger, Lamb, Miikkulainen, and Silver]{garcez2015neural}
Garcez, A., Besold, T.~R., Raedt, L., Foldiak, P., Hitzler, P., Icard, T.,
  Kuhnberger, K.-U., Lamb, L.~C., Miikkulainen, R., and Silver, D.~L.
\newblock Neural-symbolic learning and reasoning: contributions and challenges.
\newblock 2015.

\bibitem[Goodfellow et~al.(2016)Goodfellow, Bengio, Courville, and
  Bengio]{goodfellow2016deep}
Goodfellow, I., Bengio, Y., Courville, A., and Bengio, Y.
\newblock \emph{Deep learning}, volume~1.
\newblock MIT press Cambridge, 2016.

\bibitem[Gretton et~al.(2012)Gretton, Borgwardt, Rasch, Sch{\"o}lkopf, and
  Smola]{gretton2012kernel}
Gretton, A., Borgwardt, K.~M., Rasch, M.~J., Sch{\"o}lkopf, B., and Smola, A.
\newblock A kernel two-sample test.
\newblock \emph{JMLR}, 2012.

\bibitem[Guo et~al.(2017)Guo, Tang, Ye, Li, and He]{guo2017deepfm}
Guo, H., Tang, R., Ye, Y., Li, Z., and He, X.
\newblock Deepfm: a factorization-machine based neural network for ctr
  prediction.
\newblock \emph{arXiv}, 2017.

\bibitem[He et~al.(2016)He, Zhang, Ren, and Sun]{he2016deep}
He, K., Zhang, X., Ren, S., and Sun, J.
\newblock Deep residual learning for image recognition.
\newblock In \emph{CVPR}, 2016.

\bibitem[He \& Chua(2017)He and Chua]{he2017nfm}
He, X. and Chua, T.-S.
\newblock Neural factorization machines for sparse predictive analytics.
\newblock In \emph{SIGIR}, 2017.

\bibitem[Hines(1996)]{hines1996logarithmic}
Hines, J.~W.
\newblock A logarithmic neural network architecture for unbounded non-linear
  function approximation.
\newblock In \emph{ICNN}, 1996.

\bibitem[Hochreiter \& Schmidhuber(1997)Hochreiter and
  Schmidhuber]{1997Hochreiterlstm}
Hochreiter, S. and Schmidhuber, J.
\newblock {Long short-term memory}.
\newblock Number~8, pp.\  1735--1780. Neural computation, 1997.

\bibitem[Jannach et~al.(2020)Jannach, de~Souza P.~Moreira, and
  Oldridge]{jannach2020deep}
Jannach, D., de~Souza P.~Moreira, G., and Oldridge, E.
\newblock Why are deep learning models not consistently winning recommender
  systems competitions yet? a position paper.
\newblock In \emph{Proceedings of the Recommender Systems Challenge 2020}, pp.\
   44--49. 2020.

\bibitem[Kendall et~al.(1946)]{kendall1946advanced}
Kendall, M.~G. et~al.
\newblock The advanced theory of statistics.
\newblock \emph{The advanced theory of statistics.}, \penalty0 (2nd Ed), 1946.

\bibitem[Li \& Chen(2013)Li and Chen]{li2013automatic}
Li, H. and Chen, Y.~Q.
\newblock Automatic 3d reconstruction of mitochondrion with local intensity
  distribution signature and shape feature.
\newblock In \emph{ICIP}, 2013.

\bibitem[Li et~al.(2013)Li, Liu, and Chen]{li2013automatictracking}
Li, H., Liu, Y., and Chen, Y.~Q.
\newblock Automatic trajectory measurement of large numbers of crowded objects.
\newblock \emph{Optical Engineering}, 52\penalty0 (6):\penalty0 067003, 2013.

\bibitem[Li et~al.(2019)Li, Hu, Zhang, Qi, and Song]{li2019double}
Li, H., Hu, K., Zhang, S., Qi, Y., and Song, L.
\newblock Double neural counterfactual regret minimization.
\newblock In \emph{ICLR}, 2019.

\bibitem[Lian et~al.(2018)Lian, Zhou, Zhang, Chen, Xie, and
  Sun]{lian2018xdeepfm}
Lian, J., Zhou, X., Zhang, F., Chen, Z., Xie, X., and Sun, G.
\newblock xdeepfm: Combining explicit and implicit feature interactions for
  recommender systems.
\newblock In \emph{SIGKDD}, 2018.

\bibitem[Pi et~al.(2019)Pi, Bian, Zhou, Zhu, and Gai]{pi2019practice}
Pi, Q., Bian, W., Zhou, G., Zhu, X., and Gai, K.
\newblock Practice on long sequential user behavior modeling for click-through
  rate prediction.
\newblock In \emph{SIGKDD}, 2019.

\bibitem[Pi et~al.(2020)Pi, Zhou, Zhang, Wang, Ren, Fan, Zhu, and
  Gai]{pi2020search}
Pi, Q., Zhou, G., Zhang, Y., Wang, Z., Ren, L., Fan, Y., Zhu, X., and Gai, K.
\newblock Search-based user interest modeling with lifelong sequential behavior
  data for click-through rate prediction.
\newblock In \emph{CIKM}, 2020.

\bibitem[Potdar et~al.(2017)Potdar, Pardawala, and Pai]{potdar2017comparative}
Potdar, K., Pardawala, T.~S., and Pai, C.~D.
\newblock A comparative study of categorical variable encoding techniques for
  neural network classifiers.
\newblock \emph{International journal of computer applications}, 2017.

\bibitem[Qu et~al.(2020)Qu, Li, Liu, Xiong, Zhang, Chu, Wang, Qi, and
  Song]{qu2020intention}
Qu, C., Li, H., Liu, C., Xiong, J., Zhang, J., Chu, W., Wang, W., Qi, Y., and
  Song, L.
\newblock Intention propagation for multi-agent reinforcement learning.
\newblock \emph{arXiv preprint arXiv:2004.08883}, 2020.

\bibitem[Qu et~al.(2016)Qu, Cai, Ren, Zhang, Yu, Wen, and Wang]{qu2016product}
Qu, Y., Cai, H., Ren, K., Zhang, W., Yu, Y., Wen, Y., and Wang, J.
\newblock Product-based neural networks for user response prediction.
\newblock In \emph{ICDM}, 2016.

\bibitem[Radford et~al.(2019)Radford, Wu, Child, Luan, Amodei, and
  Sutskever]{radford2019language}
Radford, A., Wu, J., Child, R., Luan, D., Amodei, D., and Sutskever, I.
\newblock Language models are unsupervised multitask learners.
\newblock \emph{OpenAI blog}, 2019.

\bibitem[Rendle(2010)]{rendle2010factorization}
Rendle, S.
\newblock Factorization machines.
\newblock In \emph{ICDM}, 2010.

\bibitem[Rudolph et~al.(2016)Rudolph, Ruiz, Mandt, and
  Blei]{rudolph2016exponential}
Rudolph, M.~R., Ruiz, F.~J., Mandt, S., and Blei, D.~M.
\newblock Exponential family embeddings.
\newblock In \emph{NIPS}, 2016.

\bibitem[Saxe et~al.(2013)Saxe, McClelland, and Ganguli]{saxe2013exact}
Saxe, A.~M., McClelland, J.~L., and Ganguli, S.
\newblock Exact solutions to the nonlinear dynamics of learning in deep linear
  neural networks.
\newblock \emph{arXiv}, 2013.

\bibitem[Silver et~al.(2016)Silver, Huang, Maddison, Guez, Sifre, Driessche,
  and et~al.]{2016+silver+mastering}
Silver, D., Huang, A., Maddison, C.~J., Guez, A., Sifre, L., Driessche, G.
  V.~D., and et~al., J.~S.
\newblock {Mastering the game of Go with deep neural networks and tree search}.
\newblock \emph{Nature}, \penalty0 (7587), 2016.

\bibitem[Song et~al.(2019)Song, Shi, Xiao, Duan, Xu, Zhang, and
  Tang]{song2019autoint}
Song, W., Shi, C., Xiao, Z., Duan, Z., Xu, Y., Zhang, M., and Tang, J.
\newblock Autoint: Automatic feature interaction learning via self-attentive
  neural networks.
\newblock In \emph{CIKM}, 2019.

\bibitem[Sun et~al.(2020)Sun, Chen, Li, and Song]{sun2020improving}
Sun, H., Chen, W., Li, H., and Song, L.
\newblock Improving learning to branch via reinforcement learning.
\newblock \emph{Neural Information Processing Systems (NeurIPS) on Learning
  Meets Combinatorial Algorithms}, 2020.

\bibitem[Sutton et~al.(1998)Sutton, Barto, et~al.]{sutton1998introduction}
Sutton, R.~S., Barto, A.~G., et~al.
\newblock \emph{Introduction to reinforcement learning}, volume 135.
\newblock MIT press Cambridge, 1998.

\bibitem[Tay et~al.(2020)Tay, Dehghani, Abnar, Shen, Bahri, Pham, Rao, Yang,
  Ruder, and Metzler]{tay2020long}
Tay, Y., Dehghani, M., Abnar, S., Shen, Y., Bahri, D., Pham, P., Rao, J., Yang,
  L., Ruder, S., and Metzler, D.
\newblock Long range arena: A benchmark for efficient transformers.
\newblock \emph{arXiv preprint arXiv:2011.04006}, 2020.

\bibitem[Vaswani et~al.(2017)Vaswani, Shazeer, Parmar, Uszkoreit, Jones, Gomez,
  Kaiser, and Polosukhin]{vaswani2017attention}
Vaswani, A., Shazeer, N., Parmar, N., Uszkoreit, J., Jones, L., Gomez, A.~N.,
  Kaiser, {\L}., and Polosukhin, I.
\newblock Attention is all you need.
\newblock In \emph{NeurIPS}, 2017.

\bibitem[Wang et~al.(2021)Wang, Chang, Li, Chu, Li, Zhang, He, Song, Zhou, and
  Yang]{wang2021tcl}
Wang, L., Chang, X., Li, S., Chu, Y., Li, H., Zhang, W., He, X., Song, L.,
  Zhou, J., and Yang, H.
\newblock Tcl: Transformer-based dynamic graph modelling via contrastive
  learning.
\newblock \emph{arXiv preprint arXiv:2105.07944}, 2021.

\bibitem[Wang et~al.(2017)Wang, Fu, Fu, and Wang]{wang2017deep}
Wang, R., Fu, B., Fu, G., and Wang, M.
\newblock Deep \& cross network for ad click predictions.
\newblock In \emph{ADKDD}. 2017.

\bibitem[Xi et~al.(2020)Xi, Zhuang, Song, Zhu, Chen, Hong, Chen, Gu, and
  He]{xi2020neural}
Xi, D., Zhuang, F., Song, B., Zhu, Y., Chen, S., Hong, D., Chen, T., Gu, X.,
  and He, Q.
\newblock Neural hierarchical factorization machines for user's event sequence
  analysis.
\newblock In \emph{SIGIR}, 2020.

\bibitem[Xiao et~al.(2017)Xiao, Ye, He, Zhang, Wu, and
  Chua]{xiao2017attentional}
Xiao, J., Ye, H., He, X., Zhang, H., Wu, F., and Chua, T.-S.
\newblock Attentional factorization machines: Learning the weight of feature
  interactions via attention networks.
\newblock \emph{arXiv}, 2017.

\bibitem[Zhang et~al.(2016)Zhang, Du, and Wang]{zhang2016deep}
Zhang, W., Du, T., and Wang, J.
\newblock Deep learning over multi-field categorical data.
\newblock In \emph{European conference on information retrieval}, 2016.

\bibitem[Zhao et~al.(2017)Zhao, Li, Duan, Wang, and Chen]{zhao2017rapid}
Zhao, J., Li, H., Duan, M., Wang, S.~H., and Chen, Y.~Q.
\newblock Rapid identification of neuronal structures in electronic microscope
  image using novel combined multi-scale image features.
\newblock \emph{Neurocomputing}, 230:\penalty0 152--159, 2017.

\bibitem[Zhou et~al.(2018)Zhou, Zhu, Song, Fan, Zhu, Ma, Yan, Jin, Li, and
  Gai]{zhou2018deep}
Zhou, G., Zhu, X., Song, C., Fan, Y., Zhu, H., Ma, X., Yan, Y., Jin, J., Li,
  H., and Gai, K.
\newblock Deep interest network for click-through rate prediction.
\newblock In \emph{SIGKDD}, 2018.

\bibitem[Zhu et~al.(2020)Zhu, Xi, Song, Zhuang, Chen, Gu, and
  He]{zhu2020modeling}
Zhu, Y., Xi, D., Song, B., Zhuang, F., Chen, S., Gu, X., and He, Q.
\newblock Modeling users’ behavior sequences with hierarchical explainable
  network for cross-domain fraud detection.
\newblock In \emph{Proceedings of The Web Conference}, 2020.

\end{thebibliography}

\clearpage
\appendix

\section{Proof}\label{sec:appendixA}

\subsection{Proof of Lemma~\ref{lem:rep_trick2}}
\label{sec:proof_rep_trick2}

\begin{proof}
\begin{flalign}
\large
  \begin{aligned}
&<\boldsymbol{w_{i_1}}, \boldsymbol{w_{i_2}}, \cdots, \boldsymbol{w_{i_d}}>x_{i_1} x_{i_2} \cdots x_{i_d} \\
=&\sum^{k}_{l=1} w_{i_1l} w_{i_2l} \cdots w_{i_dl}x_{i_1} x_{i_2} \cdots x_{i_d} \\
=&\sum^{k}_{l=1} x^{v_{i_1l}}_{i_1} x^{v_{i_2l}}_{i_2} \cdots x^{v_{i_dl}}_{i_d}~~~\small{\text{According to Lemma~\ref{lem:rep_trick1}}} \\ 
=&\sum^{k}_{l=1}(x^{\boldsymbol{v_{i_1}}}_{i_1} \circ x^{\boldsymbol{v_{i_2}}}_{i_2} \circ \cdots \circ x^{\boldsymbol{v_{i_d}}}_{i_d})_l \\
  \end{aligned}
\end{flalign}

\end{proof}

\begin{restatable}{lem}{lemreptrick}
\label{lem:rep_trick1}
(1) (scalar version) For any $x, y \in \mathbb{R^{+}}$, there exists a real number $z=\frac{\ln{y}}{\ln{x}} + 1 \in \mathbb{R}$ that makes $xy=x^{z}$ hold.
Similarly, (2) (vector version) for any $x \in \mathbb{R^{+}}, \boldsymbol{y} \in (\mathbb{R}^{+})^k$, there exists a vector $\boldsymbol{z}=\frac{\ln{\boldsymbol{y}}}{\ln{x}} + 1 \in \mathbb{R}^{k}$ that makes $x\boldsymbol{y}=x^{\boldsymbol{z}}$ hold~\footnote{We conduct an element-wise operation for the vector version.}.
\end{restatable}

\subsection{Proof of Lemma~\ref{lem:rep_trick1}}
\label{sec:proof_rep_trick1}

\begin{proof}
\large
$$\ln{x^{z}} = z\ln{x} = \Big(\frac{\ln{y}}{\ln{x}} + 1 \Big) \ln{x} = \ln{x} + \ln{y} = \ln{(xy)}. $$ 
The conclusion for scalar version holds. Similarly, we conduct element-wise operation and can prove the vector version also holds.
\end{proof}

\subsection{Proof of Theorem~\ref{thm:high_order}}
\label{sec:proof_high_order}

\begin{flalign}
\large
  \begin{aligned}
g(\boldsymbol{x}) &:= \underbrace{<\boldsymbol{w^{(1)}}, \boldsymbol{x}>}_{linear (1st~order)} 
     + \underbrace{\sum_{i_1<i_2}
     \sum^{k}_{l=1}(x^{\boldsymbol{v^{(2)}_{i_1}}}_{i_1} \circ x^{\boldsymbol{v^{(2)}_{i_2}}}_{i_2})_l
     }_{2nd~order} 
     + \cdots 
     + \underbrace{\sum_{i_1<\cdots<i_d}
     \sum^{k}_{l=1}(x^{\boldsymbol{v^{(d)}_{i_1}}}_{i_1} \circ x^{\boldsymbol{v^{(d)}_{i_2}}}_{i_2} \circ \cdots \circ x^{\boldsymbol{v^{(d)}_{i_d}}}_{i_d})_l
     }_{d-th~order} \\
     &= <\boldsymbol{w^{(1)}}, \boldsymbol{x}> 
     + \sum^{k}_{l=1}\Bigg(\underbrace{\sum_{i_1<i_2}(x^{\boldsymbol{v^{(2)}_{i_1}}}_{i_1} \circ x^{\boldsymbol{v^{(2)}_{i_2}}}_{i_2})_l}_{2nd~order~embedding~summation} 
     + \cdots 
     + \underbrace{\sum_{i_1<\cdots<i_d} (x^{\boldsymbol{v^{(d)}_{i_1}}}_{i_1} \circ x^{\boldsymbol{v^{(d)}_{i_2}}}_{i_2} \circ \cdots \circ x^{\boldsymbol{v^{(d)}_{i_d}}}_{i_d})_l}_{d-th~order~embedding~summation} \Bigg) \\
     &= <\boldsymbol{w^{(1)}}, \boldsymbol{x}>  
     + \sum^{k}_{l=1}\Bigg(\underbrace{\sum_{i_1<i_2}\Big(exp\big(\sum^{2}_{j=1}\boldsymbol{v^{(2)}_{i_j}}\ln{x_{i_j}}\big)\Big)_l}_{2nd~order~embedding~summation} 
     + \cdots + \underbrace{\sum_{i_1<\cdots<i_d} \Big(exp\big(\sum^{d}_{j=1}\boldsymbol{v^{(d)}_{i_j}}\ln{x_{i_j}}\big)\Big)_l}_{d-th~order~embedding~summation} \Bigg) \\
     &= <\boldsymbol{w^{(1)}}, \boldsymbol{x}>  
     + \sum^{k}_{l=1}\Bigg(\underbrace{\sum_{o=1}^{|O|}\Big(exp\big(\sum^{d}_{i=1}\boldsymbol{u_{o_i}}\ln{x_{i}}\big)\Big)_l}_{arbitrary~order~embedding~summation} \Bigg)
  \end{aligned}
\end{flalign}

Let $u_o \in \mathbb{R}^{k \times d}$, $x \in \mathbb{R}^{d}$, $u_{oi} \in \mathbb{R}^{k}$.
Define $\boldsymbol{u_{o}} \ln{\boldsymbol{x}} = \sum^{d}_{i=1}\boldsymbol{u_{o_i}}\ln{x_{i}}$. We have

\begin{flalign}
\large
  \begin{aligned}
g(\boldsymbol{x}) = \boldsymbol{w^{(1)}}\boldsymbol{x}
     + \sum^{k}_{l=1}\Bigg(\sum_{o \in O}\Big(exp\big(\boldsymbol{u_{o}}\ln{\boldsymbol{x}}\big)\Big) \Bigg)_l
  \end{aligned}
\end{flalign}

where $O=\{i_1<i_2\} \cup \{i_1<i_2<i_3\} \cup \cdots \cup \{i_1<\cdots<i_d\}$ refers to the set of cases of all feature interactions, $i_j \in [1, d]$ and $\{\boldsymbol{u_{oi}}\}$ is the set of parameters.
Note that, the penultimate step to the last step is straightforward. For example, setting the corresponding $\boldsymbol{u_{o_i}}=\boldsymbol{0}$ for all $i>2$, equation~\eqref{eq:re_general_high_order} contains all the 2-order feature interactions.

\section{Feature Encoding}
\label{app:feature_encoding}

\begin{figure*}[h!]
\centering
\includegraphics[width=0.98\textwidth]{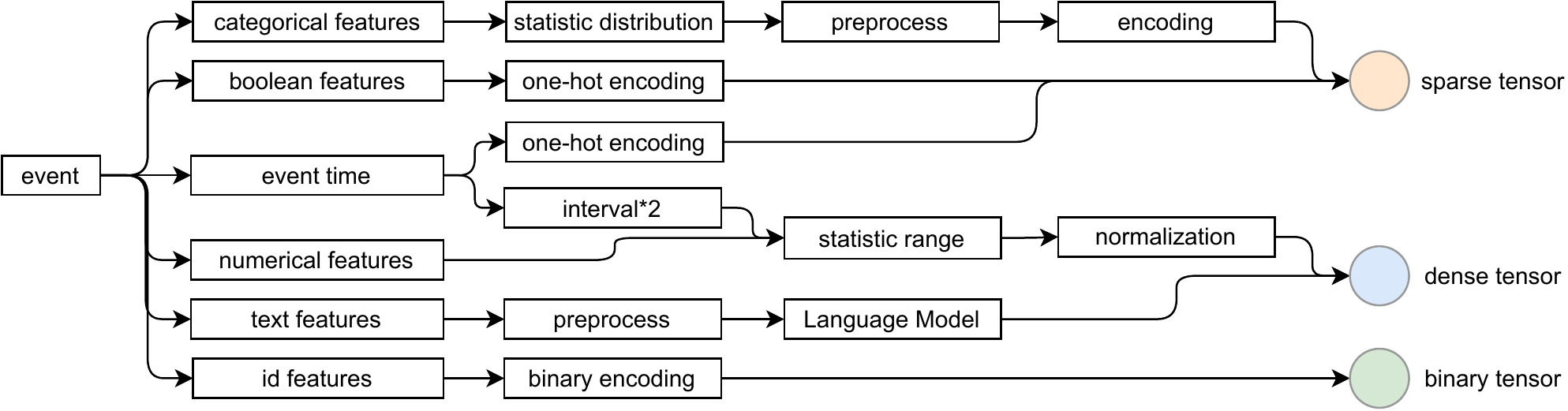}
\caption{A general approach to encode a rich set of unstructured data in the event.
}
\label{fig:data_processing}
\end{figure*}

Deep learning algorithms typically accept numerical inputs. However, industrial data in its raw form contain various data types (as illustrated in Figure~\ref{fig:data_processing}).
It's well-known that carefully pre-processed data benefit downstream deep learning task~\citep{potdar2017comparative, zhu2020modeling}.
Figure~\ref{fig:data_processing} presents a general approach to encode different types of features.
We will explain the detailed data process approach step by step as followings.

\begin{itemize}

\item \textbf{multiple categorical features.}
Typically, each event contains multiple categorical features, such as trade channel and phone types, $etc$. We will statistic their frequencies in the training samples. For the categorical variables with low frequencies~\footnote{Typically the cutoff of \textit{low} frequency is given via heuristic method according to the specific task.}, we'd like to specify them a special group named \textit{lowfreq}.
There are at least three reasons to assign \textit{lowfreq} for these categorical features:
(1) it's very difficult for the statistical learning algorithms to learn the stable pattern when only given a few numbers of data.
(2) low frequency and missing value (\textit{none} or \textit{null}) typically deliver two different information. After assigning the low-frequency categorical features the same value \textit{lowfreq}, we handle these features indistinguishably. After that, the \textit{lowfreq} features are grouped into a high-frequency feature, which could make the statistical learning algorithm much more stable.
(3) After grouping the \textit{lowfreq} categorical features, we have fewer categorical features, $i.e.$, we can use fewer parameters and memory in the downstream learning procedure.
After this important preprocessing, we can encode the categorical variables via one-hot encoding, binary encoding~\citep{potdar2017comparative} or trainable embedding~\citep{devlin2018bert}.
In our method, we select trainable embedding as the default encoding method for the categorical variable unless otherwise specified.

\item \textbf{multiple boolean features.}
Boolean feature is a special categorical variable, which only has three different values (1, 0 and \textit{none}) in the real-world data. We use the simple one-hot encoding to represent the boolean feature.

\item \textbf{event time}.
Event time is an important feature. For example, the fraudsters typically make some fraud trades late at night or make high-frequency trades within a very short time.
We can treat event time as a continuous variable (such as timestamp) or some discrete variables (such as month, weekday, hour, minute).
Although deep learning can approximate the discrete variables from their continuous values, we find it's not very efficient in practice.
Typically, when encoding the event time via expert knowledge, the deep learning method can learn some important patterns from only a few training samples.
In this paper, both continuous and discrete encoding methods are employed to represent the event time.
Specifically, one-hot encoding method transforms the event time to the discrete features as shown in Figure~\ref{fig:data_processing}.
We can make two continuous features: the timestamp interval of adjacent events and the timestamp interval between this event and the latest event. 
For the continuous features, we use \textit{min-max} method to normalize them into the range [0, 1]~\footnote{min-max normalization function: f(x)=(x-min)/(max-min).Typically, the min and max are computed according to the training data. In practice, one needs to remove the outliers from the data.}.

\item \textbf{multiple numerical features.}
In each event, there are many raw numerical features, such as trade amount, the product price and the number of products, $etc$. We will compute the range for each numerical feature and then use \textit{min-max} method to normalize these features.

\item \textbf{multiple text features.}
There are many text features in the event, such as the product name, shop name. Typically, these text features contain much useful information. 
For example, (1) buyer can obtain the style and size from the name of a product about clothes, (2) fraud merchants may carefully design their product names and make it easy for some users to access their products.
In the past years, many powerful natural language processing (NLP) techniques~\citep{devlin2018bert, radford2019language} have been proposed to learn the text presentation and human language model via self-supervised learning algorithms.
In this paper, we use the similar pre-trained language models to learn the embedding for each text feature. Note that, in the downstream learning task, these text embeddings are untrainable.

\item \textbf{multiple id features.}
When an event is generated, there are various of environmental entities, $i.e.$, multiple id features, such as device id, user id, buyer id, product id, $etc$.
One can encode these id features via one-hot encoding or trainable embedding using the similar approach as the categorical features, however, it's very expensive when there are billions of unique values for each id.
For example, when using trainable embedding with size=16 to represent $10^{10}$ ids, we need a large dictionary with size=$16 \times 10^{10}$ to save the embedding lookup table.
In this paper, we number each unique id from zero and use the binary encoding of its index as the feature.
Specifically, a vector with size=$\lceil log_{2}(N) \rceil$ is enough to represent $N$ unique ids.

\end{itemize}

\section{Baselines}
\label{sec:baseline}

We compare the proposed SHORING against state-of-the-art end-to-end feature interaction models, including:
(1) W\&D~\citep{cheng2016wide} is a deep neural network containing both wide and deep layers;
(2) DeepFM~\citep{guo2017deepfm} refers to deep factorization machines, which combines the learned embeddings of factorization machines and a deep neural network; 
(3) AFM~\citep{xiao2017attentional} refers to  attentional factorization machines and uses an attention mechanism to learn the interaction weight between different feature field;
(4) Product neural network~(PNN)~\citep{qu2016product} is another deep learning method to learn interactive patterns from multiple fields of categorical features via a product layer;
(5) HOFM~\citep{blondel2016higher} refers to high-order factorization machines and use ANOVA kernel~\citep{blondel2016polynomial} to reduce the number of parameters;
(6) DCN~\citep{wang2017deep} is deep and cross network;
(7) AFN~\citep{cheng2020adaptive} is an adaptive factorization network and used a logarithmic transformation layer to learn the arbitrary-order cross features;
(8) AutoInt~\citep{song2019autoint} used multi-head self-attention to automatically learn the feature interactions between different fields;
(9) DIN~\citep{zhou2018deep} used an attention mechanism to learn the aggregation weight for the concatenated trainable embedding.
These methods are strong enough to be used as the baselines in our experiments.

\section{Parameters and Reproducibility}
\label{sec:parameter}

Typically, the performance of deep learning model relies on its hyperparameters.
To make a fair comparison, we use the grid-search technique to find the optimal hyperparameters for each method.
We list the range of hyperparameters. The hidden dimension for event network:$\{4, 8, 16, 32\}$, $N$ in equation~\ref{eq:high_order_emd}: $\{12, 24, 48\}$, hidden dimension for sequence network and task network: $\{64, 128, 256\}$, learning rate: $\{1, 5, 10, 100\}\times 10^{-5}$, batch size: $\{64, 128, 256\}$, the number of heads $H$: $\{1, 2, 4\}$, the stack number: $2$, activation function:$\{relu\}$, $\epsilon=10^{-7}$.
The task network in our experiment is a three-layer perceptron.
We use orthogonal method~\citep{saxe2013exact} to initialize neural network's parameters and select Adam as the default optimizer. 
To balance the number of parameters and the model's generalization, we use the mechanism of parameter sharing for the event within the same sequence.
In general, we select multi-head self-attention network as the default method to learn the sequence embeddings.
Note that, we didn't share neural network's parameters across different sequences.
We implement the project in PyTorch and conduct all the experiments in a GPU machine.

\end{document}